\documentclass{article}

\usepackage[final]{neurips_2019}

\usepackage[utf8]{inputenc} 
\usepackage[T1]{fontenc}    
\usepackage{hyperref}       
\usepackage{url}            
\usepackage{booktabs}       
\usepackage{amsfonts}       
\usepackage{nicefrac}       
\usepackage{microtype}      
\usepackage{outlines}       
\usepackage{comment}
\usepackage{todonotes}
\usepackage{xcolor}

 \newcommand{\Mcomp}{MineRL Competition on Sample Efficient Reinforcement Learning } 
 \newcommand{\mcomp}{MineRL competition } 

\title{Guaranteeing Reproducibility in Deep Learning Competitions}

\author{Brandon Houghton \thanks{bhoughton@cmu.edu}   \\
  Carnegie Mellon University 
  \And
  Stephanie Milani \\
  Carnegie Mellon University
  \And
  Nicholay Topin \\
  Carnegie Mellon University 
  \And
  William Guss \\
  Carnegie Mellon University
  \And
    Katja Hofmann  \\
  Microsoft Research 
   \And
  Diego Perez-Liebana \\
  Queen Mary Univ. College of London 
   \And
   Manuela Veloso \\
   Carnegie Mellon University 
   \And
   Ruslan Salakhutdinov \\
   Carnegie Mellon University 
}

\begin{document}

\maketitle

\section{Introduction}
Democratizing access to artificial intelligence (AI) requires competitions that promote the development of sample-efficient learning, as well as ensure the reproducibility and generalizability of results. 
Sample efficiency is important because practitioners with limited compute resources cannot readily utilize algorithms that require a massive number of samples.
The complexity of these state-of-the-art methods is outpacing advancements in computation.
Moreover, as methods and domains become more specialized, learning procedures become more fragile: often undocumented modifications can inhibit reproducible results and seeds are chosen to reflect the optimal performance of a given solution~\citep{reproduce_issues}.

Because the focus of traditional research challenges is the development of new techniques in a particular field, these challenges seek to reward  participants for novel solutions. 
However, submissions with the best performance on the (often highly specified) task tend leverage domain knowledge that is not broadly applicable, leading challenges to open separate tracks where submissions are subjectively evaluated on research novelty~\citep{pavlov_mikhail_kolesnikov_sergey_m._2018}. 

To encourage participants to develop methods with reproducible and robust training behavior, we propose a challenge paradigm where competitors are evaluated directly on the performance of their learning procedures rather than pre-trained agents.
Since competition organizers re-train submissions in a controlled setting they can guarantee reproducibility, and -- by retraining submissions using a held-out test set -- help ensure generalization of submissions past the environments on which they were trained.


\section{Case Study: MineRL}
We use the aforementioned paradigm in our competition, the \Mcomp~\citep{guss_minerl_neurips2019}.
Through this competition, we challenge the deep reinforcement learning (DRL) community to train an agent to solve a complex, hierarchical task using limited computation time and a fixed budget of 8 million environment samples. 
To assist with the development of their algorithms, participants can leverage a large annotated dataset of demonstrations~\citep{guss_minerl_ijcai2019} through a competition starter kit~\citep{minerl_starterpack}.
To ensure that winning entries can be reproduced, organizers retrain submissions in the final round using an entirely new, previously-unseen texture pack~\citep{texture_pack}. 
Because the competition organizers supervise the training procedure, they can ensure that submissions hold to specific constraints (such as using a limited amount of environment samples and training time).
This requirement also prevents participants from using prior knowledge of the environment, Minecraft, to hand-craft policies. 

\subsection{Computational Requirements}
An important concern is the additional budget that this evaluation structure requires of the organizers.
In particular, organizers need additional computational power to retrain participant models. 
In order to reduce the computation required when re-training competitor submissions in the \mcomp we chose to limit re-training to round two, where the top ten teams from round one compete.
This structure allows us to open the competition to any number of interested participants while constraining the computational budget.
Additionally, by inviting only the top ten teams to a second round, we can provide each team with up to five attempts to re-train their model in round two.
We encourage organizers of future competitions to consider a similar scheme for computation allocation as a way to provide a sufficient amount of computational resources to top teams while simultaneously not limiting the number of competition participants.

\subsection{Data Requirements}

Requiring algorithms that restrict compute time and number of environment samples can result in solutions which underfit to the training data or that fail to learn even simple tasks.
One way to improve the sample efficiency of learning algorithms is to use demonstrations~\citep{dubey_humanpriors2018}. 
Many widely-used imitation learning methods require the label of the demonstration to be provided by an expert~\citep{gail_2016}; however, expert labelling is generally prohibitively expensive.
As an alternative, defining simple metrics (such as time to completion) for sub-tasks that we believed would be useful for competition, allowed us to crowd source demonstrations and provide the competitors the option to sample both expert and non-expert trajectories.

In addition to computational requirements, organizers need enough data so that both the original training set and held-out set are sufficiently large.
To meet this requirement, we recorded our dataset in such a way that it can be easily re-rendered and altered.
Specifically, to create a new dataset using the original recordings, we use these recordings to re-simulate the game actions in an environment with visual changes and capture the resulting video stream.
As a result, we create two datasets which contain the same higher-level information but which are visually distinct.
Through this process, we create two different datasets without reducing the size of either one. 

\section{Takeaways}
Through our competition, we have learned the following lessons that we would like to share with the broader community.
First, adding the constraint that all final submissions of the participants are retrained on a new texture of the environment guarantees that their submissions are reproducible and robust to perturbations. 
This requirement also limits the exploitation of domain knowledge of the environment.
Second, if organizers cannot collect a dataset rich enough for standard RL methods, they should consider potential subtasks that could be learned in a simpler domain. 
Data on these auxiliary tasks can then be given to participants to help their agents learn more basic, composable skills.
Third, some participants may become disengaged when issues impede their submission's development. 
Providing both text and video demonstration of submission gives participants confidence that their idea can be executed and encourages them to continue developing their submission.

\section{Conclusion}

We propose a novel challenge paradigm in which competitors are (1) evaluated solely on the performance of their learning procedures instead of on pre-trained agents and (2) encouraged to produce learning algorithms which prioritize sample efficiency.
Through the case study of the MineRL competition, we show that this paradigm is possible and provide examples of how to implement this paradigm in practice.
Although the proposed paradigm is computationally expensive for the competition organizers, enabling research competitions in machine learning to yield reproducible and sample-efficient methodologies provides great benefit to the community.

\subsubsection*{Acknowledgments}
We thank the following people for their contributions to the MineRL competition: Cayden Codel; Phillip Wang; Noboru (Sean) Kuno, Harm van Seijen, Andre Kramer, and Microsoft Research; Sharada Mohanty, Shivam Khandelwal, Shinya Shiroshita, and AICrowd; Crissman Loomis, Mario Ynocente Castro, Keisuke Nakata, Shohe Hido, and Preferred Networks; Chelsea Finn; Oriol Vinyals; and Sergey Levine. 


\small
\bibliographystyle{plainnat}

\end{document}